\documentclass[sigconf,nonacm]{acmart}

\usepackage{times}
\usepackage{latexsym}

\usepackage{url}
\usepackage[normalem]{ulem} 
\usepackage{soul} 
\usepackage{graphicx}
\usepackage{amsmath}
\usepackage{subcaption} 
\usepackage{array}
\usepackage{multirow}

\AtBeginDocument{%
  \providecommand\BibTeX{{%
    \normalfont B\kern-0.5em{\scshape i\kern-0.25em b}\kern-0.8em\TeX}}}

\usepackage{svg}
\newcommand{\orcidPNG}[1]{\href{https://orcid.org/#1}{\includegraphics[width=10pt]{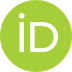}}}

\begin{document}

\title{Nested and Balanced Entity Recognition \\ using Multi-Task Learning}

\author{Andreas Waldis \orcidPNG{0000-0002-2772-5701}}
\email{andreas.waldis@hslu.ch}
\email{andreas.waldis@live.com}
\orcid{0000-0002-2772-5701}
\affiliation{%
  \institution{HSLU - Lucerne University of Applied Sciences and Arts\\School of Computer Science and Information Technology\\Information Systems Lab}
  \streetaddress{Suurstoffi 1}
  \city{CH-6343, Rotkreuz}
  \country{Switzerland}
}

\author{Luca Mazzola \orcidPNG{0000-0002-6747-1021}}
\email{luca.mazzola@hslu.ch} 
\email{mazzola.luca@gmail.com}
\orcid{0000-0002-6747-1021}
\affiliation{%
  \institution{HSLU - Lucerne University of Applied Sciences and Arts\\School of Computer Science and Information Technology\\Information Systems Lab}
  \streetaddress{Suurstoffi 1}
  \city{CH-6343, Rotkreuz}
  \country{Switzerland}
}

\renewcommand{\shortauthors}{Waldis and Mazzola}


\begin{abstract}
Entity Recognition (ER) within a text is a fundamental exercise in Natural Language Processing, enabling further depending tasks such as Knowledge Extraction, Text Summarisation, or Keyphrase Extraction.
An entity consists of single words or of a consecutive sequence of terms, constituting the basic building blocks for communication.
Mainstream ER approaches are mainly limited to flat structures, concentrating on the outermost entities while ignoring the inner ones. 
This paper introduces a partly-layered network architecture that deals with the complexity of overlapping and nested cases.
The proposed architecture consists of two parts: (1) a shared Sequence Layer and (2) a stacked component with multiple Tagging Layers.
The adoption of such an architecture has the advantage of preventing overfit to a specific word-length, thus maintaining performance for longer entities despite their lower frequency.
To verify the proposed architecture's effectiveness, we train and evaluate this architecture to recognise two kinds of entities - Concepts (CR) and Named Entities (NER).
Our approach achieves state-of-the-art NER performances, while it outperforms previous CR approaches.
Considering these promising results, we see the possibility to evolve the architecture for other cases such as the extraction of events or the detection of argumentative components.  

\end{abstract}


\ccsdesc[500]{Information systems~Language models}
\ccsdesc[500]{Computing methodologies~Natural language processing}
\ccsdesc[500]{Computing methodologies~Information extraction}
\ccsdesc[500]{Computing methodologies~Neural networks}

\keywords{Named Entity Recognition, Concept Recognition, Multi-Task Learning, Natural Language Processing}


\maketitle
\balance 
\pagestyle{plain}

\section{Introduction}

Natural Language Processing aims to handle and analyse many data in an automatic and computer-based manner.
Other tasks - such as understanding text or interacting with humans - can build on this resulting knowledge.
One fundamental task of text understanding is to find entities - as a word or word combinations - within a given fragment.
If one thinks about a text as a carrier of information, an author use these objects as basics building blocks to encode the idea and transmit it to others.
Thus, the kind of these entities differs based on their purpose or topic.
For example, items that capture sentiments usually convey the message using other words and following other structures compared to ones describing real-world objects.

\begin{figure}[ht]
\centering
\includegraphics[width=1.0\linewidth]{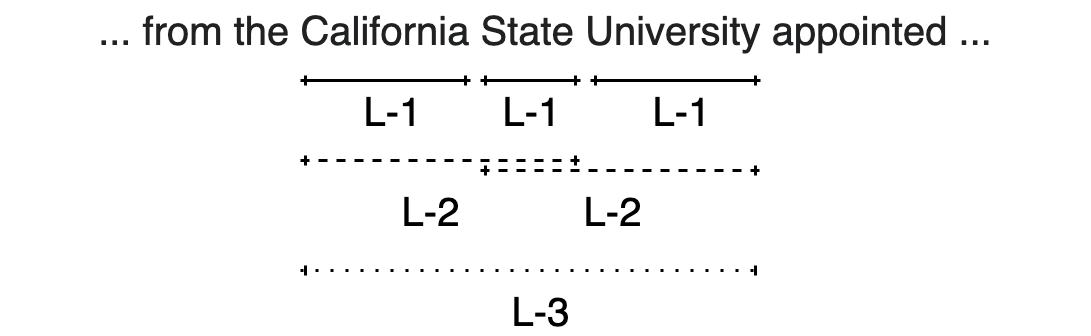}
\caption{Example with overlapping entities and three nested levels - level-1 with three entities of the length one (L-1, solid line), level-2 with two entities with two words (L-2, dashed line), and level-3 with one entity of length three (L-3, dotted line)}
\label{fig:sample-sentence}
\end{figure}

While humans can naturally detect named or un-named entities and understand their differences, a computer algorithm struggles to effectively achieve the same result.
Two reasons for these difficulties are the fact that entities can overlap each other and their nested occurrences.
As the sample "California State University" in \autoref{fig:sample-sentence} shows, 
there are overlapping cases like the two entities "California State" and "State University".
Likewise, there are the nested entities like "California", "State", "University", "California State", and "State University" within "California State University".
In this sample, there are three levels of such nested entities. From the innermost level one (L-1) which does not embed samples of lower nested-levels to the outermost level three (L-3), including samples of level two (L-2) and one. 

To address these complexities, we introduce in the following work a partly-layered sequence-based neural network for Entity Recognition.
This approach shapes this recognition as a token classification task. Thereby, it can consider overlapping and nested cases - as in \autoref{fig:sample-sentence}.
To evaluate this network architecture's effectiveness, we use it to recognise two types of entities - concepts and named entities.
In contrast to other work (\cite{campos2013modular,tseytlin2016noble}), we differentiate between concepts, as words and word combinations with a high information density~\cite{parameswaran2010towards}, and named entities, as a real-world object like a person, an organisation, or a location.
While there is a higher agreement among scholars on what constitutes a valid named entity, this is not the case for concepts identification. 
Every human has its thoughts, whether a word combination is a concept or not. It is hard to have an objective and externalised metric to measure their validity.
Consequently, it is hard to develop generally accepted and large enough manually annotated golden standard~\cite{nothman2009analysing}.

Due to this uncertainty, every scholar or research group develops its reference data set. In our case, we follow the suggested approach of Parameswaran et al. \cite{parameswaran2010towards} and adopt the list of Wikipedia article titles as reference for valid concepts (truth base).
To remove unwanted titles, we apply - as they suggest - common-knowledge wisdom and empirical, experience-based rules. Those heuristics allow to filter the full set of Wikipedia titles by removing the ones containing no noun, or that starts or ends with a verb, conjunction, article or pronoun.
The idea behind using Wikipedia titles is to obtain a model that generalise towards most important concepts and not to produce a perfect model for identifying Wikipedia titles in a text.

To sum up, this paper main contribution aims to show how our specialised network architecture handles nested and overlapping cases. 
Further, we treat with its' layered architecture the problem of implicit imbalance within the data. 
This imbalance origins in the rarity of labels compared to the number of candidates.
As a proof of concept, this work reports an extensive evaluation for models instantiating the proposed architecture on two different tasks: Concept Recognition (CR) and Named Entity Recognition (NER). 
With a short text example, we analyse the performance and the major error types in detail for both tasks.

The rest of the paper is organised as follows: Section 2 presents the related works. It stresses the innovative aspects of our approach. The architecture is then introduced and detailed in Section 3. At the same time, Section 4 introduces the reader to discover the most promising architecture instances, the used evaluation metrics and the training settings adopted.
Eventually, this section closes by covering the set of experiments performed for nested CR and NER, including a use case to study individual predictions for both tasks. Section 5 showcases the results obtained in the different tasks and on the use case. Subsequently, Section 6 concludes the paper with a discussion of the achieved performances and the future steps we would like to take.
\section{Related Work}

Concept Recognition (CR) and Named Entity Recognition (NER) assign a particular label to a consecutive sequence of words. 
This is a binary label for CR (concept or not) \cite{bhole2007mining}, but list of predefined named entities \cite{dong2016multiclass} (person, company, etc.) in NER.
Other examples of similar tasks are argument component detection~\cite{lippi2016argumentation} or document classification~\cite{yang2016hierarchical}, where the classified sequence corresponds respectively to a sentence or the whole text.

CR is classically designed in the form of a direct sequence classification as in
~\cite{parameswaran2010towards,liu2016domain,waldis2020towards_XAI_concepts_recognition}.
Thus, a model assigns a final label to one n-gram. 
On the other hand, NER approaches encode the actual label (be it a Person, a Company, or a DNA) for every word using a BIO (Begin-, Inside-, Outside-of-an-entity) encoding schema. 
In this schema, every word gets a label as the begin, an inner element or outside of an entity. 
For example, B-PER indicates the beginning of a person entity. 
Examples of approaches for named entities identification are presented in~\cite{akbik2018contextual,li2019unified,jie2019dependency,luo2020hierarchical,kruengkrai2020improving,hu2020leveraging}.

This schema works well for a flat sequence classification when every word belongs precisely to a single class. 
However, one word can be part of more than one classes.
Such as, when two concepts overlap each other, or a named entity contains other entities with fewer words (as in "\textit{California State University}" from \autoref{fig:sample-sentence}).
For example, it could be the start of a person name (B-PER) and within a company (I-COM) at the same time. 
One can think for example of \textit{Hewlett Packard}, where \textit{Packard} refers to the person David Packard. However, it also constitutes part of the name of the company he founded. 

This fact leads to a significantly increasing number of class combinations when considering concepts and named entities with a higher number of nested levels.
Recent approaches encode these levels with a specific tagging schema~\cite{strakova2019neural}, or using special architectures like a ensemble of networks~\cite{zheng2019boundary}, a layered network~\cite{ju2018neural}, anchor-region networks~\cite{lin2019sequence}, or graph-based networks~\cite{luo2020bipartite}.

The class imbalance within the data is another aspect to treat when it comes to CR or NER.
Most of the samples do not belong to any specific class (non-concept or non-named-entity).
As showed in~\cite{li2019dice}, this imbalance can rise to a ratio of 168:1. 
One option to handle this imbalance is to over or under-sample the data
~\cite{waldis2018concept}, 
or selecting features based on their importance for the minority class 
~\cite{waldis2020towards_XAI_concepts_recognition}.
Other approaches to overcoming the negative consequences of these extreme imbalanced classes 
are to use different loss functions~\cite{li2019dice} or to adopt different weights for the distinct classes~\cite{nguyen2020adaptive}.

Although recent approaches pushed over nested NER and CR performance, we observe that almost all scientific publications report only overall performance or the per-class performance, but not an integrated view. 
We want to focus on the performance for all different word-lengths with our work, considering the imbalance within the data set.
To this objective, we propose an architecture applicable for CR and NER and regularise towards balanced learning of all word-lengths using structural and functional facets.
We use dedicated parts for each word-length as well as shared ones within the network architecture structural design.
Further, we reshape the learning process as multi-task learning. 
This setting allows us to consider the prediction of one word-length and apply separated loss functions and optimisers.
This kind of learning can help to find an optimal fit in all parts (shared \& dedicated) of a network~\cite{zhang2017survey}. 

\section{Partly-Layered Neural Network}

\begin{figure*}[htb]
\centering
\includegraphics[width=0.8\textwidth]{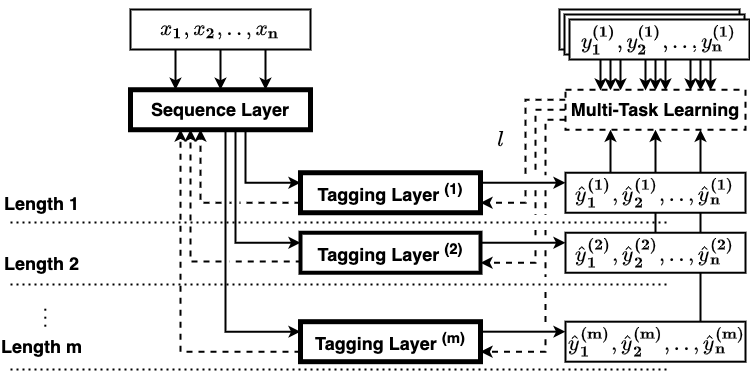}
\caption{Basic structure of the proposed approach including the structural facets (bold) and the functional facets (dashed)}
\label{fig:basic_network_structure}
\end{figure*}

The following section introduces the details of our proposed network architecture and its details.
\autoref{fig:basic_network_structure} shows a system overview and outlines its structural (Sequence \& Tagging Layer(s)) and functional facets (Multi-Task Learning). 
Our approach takes as input a given text sequence $x$ and produces a prediction sequence $\hat{y}$ for every word-length $m$.
We use the \textbf{BO} (begin, other) schema to encode the target ($y$) concepts or named entities and predictions ($\hat{y}$) as a sequence.
This schema tags every token of a sequence as begin \textbf{B} or not begin \textbf{O} of a concept or named entity.
Since the networks predict $\hat{y}$ for every word-length, we can compose the final output by concatenating a word tagged as \textbf{B} with the next $m-1$ words.

\subsection{Structural Facets}
The structural facets control the shape of the network and thereby the forward-propagation process.
As already seen, we divided it into two parts: (1) one Sequence Layer, and (2) a Tagging Layer for each word-length $m$.
The basic intuition behind this structure is that the Sequence Layer extracts necessary information about the input sequence $x$.
Afterwards, the Tagging Layers transform that into $m$ prediction sequences $\hat{y}$.
Thus, every Tagging Layer learns what is essential for the specific word-length. In contrast, the Sequence Layers learns in a way optimal for every length.
This implicit influence of all Tagging Layers allows higher word-lengths, with a higher sample rarity, to profit from lower ones.

\subsection{Functional Facets}
The functional facets control the learning process, including calculating the loss, the back-propagation, and updating the network weights using an adaptive optimiser. 
To let the Tagging Layers the full freedom to fit their weight, we split the learning process into multiple separated tasks.
Every learning task calculates the loss for one prediction sequence, back-propagates it, updates the Sequence Layer's weights, and the corresponding Tagging Layer (visualised with dash lines in~\autoref{fig:basic_network_structure}).
As loss function, we use the cross-entropy and as optimiser the AdamW~\cite{loshchilov2017fixing}.
We include a weight adjustment of the different classes within the loss function to stimulate the minority class and fight the imbalance. 
\section{Experiment}
This chapter outlines the different network variations, parametrisation, and used datasets.
To verify the proposed approach's effectiveness, we ran two quantitative experiments and a concluding qualitative case study. 
While the former focus on measuring the CR and biological NER performance, the latter focuses on the qualitative analysis of differences between the various models, where they succeed, and where they fail.

\subsection*{Architectures Variations}
\autoref{tab:used_networks} lists the seven different variation used for Concept Recognition and Named Entity Recognition - three BERT-based (\textsc{[Bert]}, \textsc{[DistilBert]}, \textsc{[Ro\-Berta]}) and four LSTM-based ones. 
The \textsc{[Mu\-lti]} variation uses a dedicated Sequence Layer for each word-length to examine the differences between a fully- and partly-layered architecture.
The \textsc{[Norm-Flair]} variations is a special case, that use combination of general (glove \cite{pennington2014glove}) and medical word embeddings (flair \cite{akbik2018coling}). 

\begin{table}
\centering
\caption{Overview of used architectures variations. }
\begin{tabular}{p{2.5cm}p{2.35cm}p{2.5cm}}
\hline
\textbf{Name} &  \textbf{Sequence Layer} & \textbf{Tagging Layer}\\ \hline \hline
\textsc{[Base]} & LSTM & Dropout, Dense \\ \hline
\textsc{[Input-Drop]} & Dropout, LSTM & Dropout, Dense \\ \hline
\textsc{[Norm]} & LSTM, Normalization & Dropout, Dense \\ \hline
\textsc{[Norm-Flair]}* & LSTM, Normalization & Dropout, 2xDense \\ \hline
\textsc{[Multi]} & Multi-LSTM & Dropout, Dense \\ \hline
\textsc{[Bert]}& Bert & Dropout, Dense \\ \hline
\textsc{[DistilBert]} & DistilBert & Dropout, Dense \\ \hline
\textsc{[RoBerta]} & RoBerta & Dropout, Dense \\ \hline\hline
\end{tabular}
\center{* indicates specific settings only adopted for NER.}
\label{tab:used_networks}
\end{table}

\subsubsection*{Metrics}

We evaluate the two experiments' performance considering the overall, the word-length, and the nested-level performance.
For all these three measures, we use the precision \textbf{P}, recall \textbf{R}, and f1-score \textbf{F1}. 
To measure CR's overall performance, we use the micro \textit{mi} and macro \textit{ma} average. At the same time, for NER, we calculate the performance overall and on every named entity. 
The word-length evaluation considers the performance for the every word length (CR), and for every combination of word length and named entity class (NER). 
For the evaluation for the different nested-levels, we measure the performance grouping the samples by their specific nested-level.

\subsection*{Training Settings}
We evaluate a broad set of hyper- and training-parameters for CR and NER - as shown in~\autoref{tab:hyperparameters-range}.
Based on the performance difference, we select the final parameters - see~\autoref{tab:hyperparameters}.
In addition,~\autoref{tab:hyperparameters-weights} lists the used class weights to calculate the loss function.
We save the network-weights after each epoch during the training process, if the macro average of the \textbf{F1} on the validation set is higher than in earlier epochs.

\begin{table*}
\centering
\caption{Selected parameters to train LSTM- and BERT-based models on \textbf{CR} and \textbf{NER} (* parameter for \textsc{[Norm-Flair]})}
\begin{tabular}{@{\extracolsep{4pt}}>{\centering\arraybackslash}p{3cm}>{\centering\arraybackslash}p{2cm}>{\centering\arraybackslash}p{2cm}>{\centering\arraybackslash}p{2cm}>{\centering\arraybackslash}p{2cm}}
 \hline
\multirow{2}{*}{\textbf{Parameter}} &\multicolumn{2}{c}{\textbf{CR}}  & \multicolumn{2}{c}{\textbf{NER}}\\ \cline{2-3} \cline{4-5}
& \textbf{LSTM} & \textbf{Bert} &  \textbf{LSTM} & \textbf{Bert}\\  \hline \hline 
Epochs& 30 & 20 & 140 & 100\\ \hline 
Batch size & 20'000 & 2'500 & 20'000, 10'000* & 20'000\\ \hline 
Embedding & glove-300d & - & glove-300d, pubmed* & -\\ \hline 
LSTM layers & 1 & - & 2 & -\\ \hline 
LSTM dropout & 0.4& - & 0.4 & -\\ \hline 
LSTM hiden dim. & 500& - & 500, 1000* & -\\ \hline 
Tagging-Layer dropout & 0.4 & 0.4 & 0.4\\ \hline 
Learning rate & 0.001 & 0.00001 & 0.001& 0.00005\\ \hline 
Input dropout & 0.2 & - & 0.2  & -\\ \hline 
\end{tabular}
\label{tab:hyperparameters}
\end{table*}

\begin{table*}
\centering
\caption{Range of parameters used for the training of LSTM- and BERT-based models for both experiments (\textbf{CR} and \textbf{NER})}
\begin{tabular}{@{\extracolsep{4pt}}>{\centering\arraybackslash}p{3cm}>{\centering\arraybackslash}p{2cm}>{\centering\arraybackslash}p{2cm}>{\centering\arraybackslash}p{2cm}>{\centering\arraybackslash}p{2cm}}
 \hline
\multirow{2}{*}{\textbf{Parameter}} &\multicolumn{2}{c}{\textbf{CR}}  & \multicolumn{2}{c}{\textbf{NER}}\\ \cline{2-3} \cline{4-5}
& \textbf{LSTM} & \textbf{Bert} &  \textbf{LSTM} & \textbf{Bert}\\  \hline \hline 
Epochs& 30 & 20 & 140 & 100\\ \hline 
Batch size & 5'000-20'000 & 2'500-10'000 & 5'000-20'000 & 2'500-20'000\\ \hline 
Embedding & glove 50d, glove 100d, glove 200d, glove 300d& - & glove 50d, glove 100d, glove 200d, glove 300d & -\\ \hline 
LSTM layers & 1-4 & - & 2 & -\\ \hline 
LSTM dropout & 0.3-0.5 & - & 0.4 & -\\ \hline 
LSTM hiden dim. & 300-600& - & 500 & -\\ \hline 
Tagging-Layer dropout & 0.3-0.5 & 0.3-0.5  & 0.3-0.5 \\ \hline 
Learning rate & 0.01-0.0001 & 0.0001-0.0000001& 0.001& 0.0001-0.0000001\\ \hline 
Input dropout & 0.1-0.3 & - & 0.2  & -\\ \hline 
\end{tabular}
\label{tab:hyperparameters-range}
\end{table*}

\begin{table*}
\centering
\caption{Used class-weights for different lengths \textbf{L} to calculate the loss: concept (\textbf{C}), non-concept (\textbf{N}), protein (\textbf{P}), DNA (\textbf{D}), RNA (\textbf{R}), cell line (\textbf{CL}), cell type (\textbf{CT})}
\begin{tabular}{@{\extracolsep{4pt}}>{\centering\arraybackslash}p{0.5cm}>{\centering\arraybackslash}p{1cm}>{\centering\arraybackslash}p{0.55cm}>{\centering\arraybackslash}p{0.7cm}>{\centering\arraybackslash}p{0.55cm}>{\centering\arraybackslash}p{0.55cm}>{\centering\arraybackslash}p{0.55cm}>{\centering\arraybackslash}p{0.55cm}>{\centering\arraybackslash}p{0.55cm}>{\centering\arraybackslash}p{0.7cm}>{\centering\arraybackslash}p{0.55cm}>{\centering\arraybackslash}p{0.55cm}>{\centering\arraybackslash}p{0.55cm}>{\centering\arraybackslash}p{0.55cm}>{\centering\arraybackslash}p{0.55cm}}
 \hline
\multirow{2}{*}{\textbf{L}} &\multicolumn{2}{c}{\textbf{CR}}&\multicolumn{6}{c}{\textbf{NER}} &\multicolumn{6}{c}{\textbf{NER-Flair}} \\ \cline{2-3} \cline{4-9} \cline{10-15} 
& \textbf{C} & \textbf{N} & \textbf{N}& \textbf{P} & \textbf{D} & \textbf{R} & \textbf{CL} & \textbf{CT}& \textbf{N}& \textbf{P} & \textbf{D} & \textbf{R} & \textbf{CL} & \textbf{CT} \\  \hline \hline 
1 & $1-2^{-2}$ & $2^{-2}$ & 0.005&	0.20&	0.20&	0.30&	0.24 & 0.21 & 0.040&	0.15&	0.18&	0.25&	0.22&	0.20\\\hline 
2 & $1-2^{-3}$ & $2^{-3}$ & 0.005&	0.20&	0.20&	0.30&	0.24 & 0.21 & 0.030&	0.15&	0.18&	0.25&	0.22&	0.20\\ \hline 
3 & $1-2^{-4}$ & $2^{-4}$ & 0.005&	0.20&	0.20&	0.30&	0.24 & 0.21 & 0.015&	0.15&	0.18&	0.25&	0.22&	0.20\\ \hline 
4 & $1-2^{-5}$ & $2^{-5}$ & 0.005&	0.20&	0.20&	0.30&	0.24 & 0.21 & 0.010&	0.15&	0.18&	0.25&	0.22&	0.20\\ \hline 
5 & $1-2^{-6}$ & $2^{-6}$ & 0.005&	0.20&	0.20&	0.30&	0.24 & 0.21 & 0.008&	0.15&	0.18&	0.25&	0.22&	0.20\\ \hline 
6 & $1-2^{-7}$ & $2^{-7}$ & 0.005&	0.20&	0.20&	0.30&	0.24 & 0.21 & 0.006&	0.15&	0.18&	0.25&	0.22&	0.20\\ \hline 
7 & $1-2^{-8}$ & $2^{-9}$ & -& -& -& -& -& -& -& -& -& -& -& -\\ \hline \hline 
\end{tabular}
\label{tab:hyperparameters-weights}
\end{table*}

\subsection{Experiment I: Concept Recognition}
The first experiment uses our architecture to recognise concepts with a word length of one to seven words - thus using seven Tagging Layers.
The training and validation dataset consists of 300'000 and 150'000 sentences from the \textbf{all-the-news} dataset\footnote{available \href{https://www.kaggle.com/snapcrack/all-the-news}{https://www.kaggle.com/snapcrack/all-the-news}} -  
a dataset that consists of 150'000 news articles from various publishers.
For testing, we use all the sentence from a hold out set of 1000 news articles 
\cite{news-samples}, 
1000 random sampled Wikipedia articles 
\cite{wiki-samples}, 
and all the documents from the full \textbf{DUC2001} dataset~\footnote{available \href{https://www-nlpir.nist.gov/projects/duc/guidelines/2001.html}{https://www-nlpir.nist.gov/projects/duc/guidelines/2001.html}} as well as the \textbf{Hulth2003} dataset \cite{hulth2003improved}. 
\autoref{tab:portion-concepts} shows the number of concepts in total, for the five nested levels, for each word length, and over the whole test set.
Besides, it shows the imbalance with regards to the number of concepts for different word-lengths and nested-levels.

\begin{table}[ht]
\centering
\caption{CR dataset insights per word-length }
\begin{tabular}{@{\extracolsep{1pt}}>{\centering\arraybackslash}p{1cm}>{\centering\arraybackslash}p{1.35cm}>{\centering\arraybackslash}p{0.9cm}>{\centering\arraybackslash}p{0.8cm}>{\centering\arraybackslash}p{0.55cm}>{\centering\arraybackslash}p{0.45cm}>{\centering\arraybackslash}p{0.35cm}}
\hline
\multirow{2}{*}{\textbf{Length}} & \multirow{2}{*}{\textbf{Concepts}} & \multicolumn{5}{c}{\textbf{Nested Level}} \\ \cline{3-7}
&  & 1th & 2nd & 3rd & 4th & 5th \\ \hline\hline
1 & 201'240  & 201'240 & - & - & - & - \\ \hline
2 & 36'630 & 3'696 & 32'934 & - & - & - \\ \hline
3 & 4'939 & 202 & 2'301 & 2436 & - & - \\ \hline
4 & 1'115 & 37 & 369 & 546 & 163 & - \\ \hline
5 & 271 & 1 & 58 & 128 & 78 & 6 \\ \hline
6 & 79 & - & 13 & 32 & 28 & 6 \\ \hline
7 & 32 & - & 4 & 15 & 13 & - \\ \hline\hline
Total  & 383'428  & 205'176 & 35'679 & 3'157 & 282 & 12\\ \hline\hline
\end{tabular}
\label{tab:portion-concepts}
\end{table}

\subsection{Experiment II: Named Entity Recognition}
In the second experiment, we recognise nested named entities.
For this purpose, we selected the commonly used GENIA Dataset \cite{kim2003genia}, consisting of biomedical named entities with a word-length of one to six. 
In contrast to other often used dataset (like ACE2004 \cite{doddington2004automatic} and ACE2005 \cite{walker2006ace}), this one is freely available.
We used the same train-, dev-, test-split (81\%, 9\%, 10\%) as other approaches (\cite{finkel2009nested,ju2018neural,li2019unified}) to make the results comparable.  
Further, we grouped the 36 sub-categories as labels \textbf{Protein}, \textbf{DNA}, \textbf{RNA}, \textbf{Cell Type}, and \textbf{Cell Line} as in \cite{finkel2009nested}.
\autoref{tab:portion-ner} shows the portion of these labels of the test set and the three nested levels.

\begin{table}[ht]
\centering
\caption{Insights of the GENIA testset per word-length}
\begin{tabular}{@{\extracolsep{1pt}}>{\centering\arraybackslash}p{1cm}>{\centering\arraybackslash}p{0.8cm}>{\centering\arraybackslash}p{0.45cm}>{\centering\arraybackslash}p{0.45cm}>{\centering\arraybackslash}p{0.45cm}>{\centering\arraybackslash}>{\centering\arraybackslash}p{0.45cm}>{\centering\arraybackslash}p{0.45cm}>{\centering\arraybackslash}p{0.35cm}>{\centering\arraybackslash}p{0.35cm}}
\hline
\multirow{2}{*}{\textbf{Length}}&  \multicolumn{5}{c}{\textbf{Named Entity}} &  \multicolumn{3}{c}{\textbf{Nested Level}} \\ \cline{2-6}\cline{7-9}
 & \textbf{Protein} & \textbf{DNA} & \textbf{RNA} & \textbf{Cell Line} & \textbf{Cell Type} & \textbf{1st} & \textbf{2nd} & 
\textbf{3rd}\\ \hline \hline
1 & 1'534 & 333 & 18 & 83 & 156 & 2'124 & -
 & - \\ \hline
2 & 836 & 461 & 71 & 144 & 298 & 1'616 & 194 & - \\ \hline
3 & 454 & 246 & 8 & 102 & 117 &	764 & 162 & 1 \\ \hline
4 & 165 & 126 & 12 & 68 & 30 & 297 & 94 & 10\\ \hline
5 & 52 & 45 & 3 & 28 & 8 & 107 & 24 & 5\\ \hline
6 & 21 & 29 & 1 & 24 & 4 & 65 & 14& -\\ \hline\hline
- & 3'062 & 1'240 & 113 & 449 & 613 & 4'973 & 488 & 16\\ \hline\hline
\end{tabular}
\label{tab:portion-ner}
\end{table}

\subsection{Experiment III: Case Study}
The concluding experiment consists of a case study focusing on examining the differences between propose model variations and identifying cases where they succeed and fail. 
This study investigates the prediction of one sentence for Concept Recognition and of another one for Named Entity Recognition.

\subsubsection{Concept Recognition}
For an in-depth analysis of resulting concepts, we use the following treacherous sentence.
It speaks about the government of the \textit{Haida Nation} located on the border of Canada and Alaska. 
We choose this sentence for two reasons:
First, it contains various overlapping and nested occurrences of potential concepts.
Second, it includes several out-of-vocabulary words like \textit{XaaydaGa}.
Thus, this example helps to study the generalisation of the different models towards unknown words.
Thereby, it allows to verify the initial requirement that a model does not overfit with regards to Wikipedia titles.
\newline
\begin{quote}
    \textit{XaaydaGa Waadluxan Naay is the \underline{elected \underline{government}} and the \underline{\underline{Council} of the \underline{Federation}} of \underline{Haida}.}
    \newline
\end{quote}

\subsubsection{Named Entity Recognition}
To analyse the predictions of the different NER models in detail, we used the following sentence. 
It covers the protein \textit{TCF-1 alpha} and its appearance in different DNA sequences like \textit{TCR delta}, and \textit{TCR beta}.
This sentence includes various nested entities of different types, and thereby provides an interesting example to test the nested recognition.
\newline
\begin{quote}
\begin{center}
    \textit{Sequences related to the $$\underline{\underline{\text{TCF-1 alpha}}_{\text{ }\textbf{Protein}} \text{binding motif}}_{\text{ }\textbf{DNA}}$$ (5'-GGCACCCTTTGA-3') are also found in the $$\underline{human \text{ } \underline{\underline{TCR}_{\text{ }\textbf{Protein}} \text{ } delta}_{\text{ }\textbf{DNA}}}_{\textbf{\text{ }DNA}}$$ (and possibly  $\underline{\underline{TCR}_{\text{ }\textbf{Protein}} beta}_{\text{ }\textbf{DNA}})\text{ } \underline{enhancers.}_{\text{ }\textbf{DNA}}$}
\newline
\end{center}
\end{quote}

\section{Results}
The following section states the quantitative results of CR and NER, followed by the predictions over the qualitative case study.

\subsection{Concept Recognition}
For CR, we compare the different network variations with previous approaches by measuring and comparing the overall performance, per word-length, and per nested-level.
Despite \cite{parameswaran2010towards}, there is - to the best of our knowledge - no other approach that treats concepts in the same way as we do in this experiment. 
Due to the different design of the experiments and the their focus on precision in \cite{parameswaran2010towards}, we select two approaches of our previous work \textbf{Method-1} \cite{waldis2020towards_XAI_concepts_recognition} and \textbf{Method-2} \cite{waldis2018concept_extraction_CNN} to compare.
While the first one uses a wide range of calculated features, the second uses Convolutional Neural Networks with different horizontal and vertical filters.

\subsubsection{Overall Performance}

\autoref{tab:concept-recognition-results} shows the overall results on the test set of the trained models and the two approaches from the literature.
The \textbf{Method-1} model achieves the highest \textbf{P} micro score (89.4\%), while \textsc{[Bert]} the highest \textbf{F1} and \textbf{P} macro score (53.5\%, 62.4\%), \textsc{[InputDrop]} the best \textbf{R} macro and micro score (80.1\%, 95.1\%), and \textsc{[Multi]} the highest \textbf{F1} micro score (89.8\%).
Overall, our best performing models outperform the best previous approach with regards to the macro measurement by 6.8-26.2\% (\textbf{F1}), 3.1-17.5\% (\textbf{P}), and 12.5-16.8\% (\textbf{R}).
This cap shrinks to 0.9-1.2\% (\textbf{F1}), and 3-8\% (\textbf{R}) when considering the micro average.
Comparing the different types of our proposed models reveals that the BERT-based outperforms the LSTM-based ones for all metrics, along with a higher average word-length per concept. 

\begin{table}
\centering
\caption{Overall \textbf{CR} performance}
\begin{tabular}{@{\extracolsep{1pt}}>{\centering\arraybackslash}p{2.1cm}>{\centering\arraybackslash}p{0.4cm}>{\centering\arraybackslash}p{0.4cm}>{\centering\arraybackslash}p{0.4cm}>{\centering\arraybackslash}p{0.4cm}>{\centering\arraybackslash}p{0.4cm}>{\centering\arraybackslash}p{0.4cm}>{\centering\arraybackslash}p{1.1cm}}
\hline
 \multirow{2}{*}{\textbf{Model}}&  \multicolumn{2}{c}{\textbf{P}} &  \multicolumn{2}{c}{\textbf{R}} &  \multicolumn{2}{c}{\textbf{F1}} & \multirow{2}{*}{\textbf{Avg. Len.}}\\ \cline{2-3}\cline{4-5}\cline{6-7}
 & \textit{ma}  & \textit{mi} & \textit{ma}  & \textit{mi} & \textit{ma}  & \textit{mi}& \\ \hline \hline

\textsc{[Base]} & 41.1 & 80.4 & 76.6 & 94.2 & 51.6 & 86.7 & 1.94\\ \hline
\textsc{[Input-Drop]} & 34.6 & 75.3 & \textbf{80.1} & \textbf{95.1} & 45.1 & 84.1 & 1.99\\ \hline
\textsc{[Multi]}  & 23.1 & 89.3 & 25.2 & 90.4 & 24.0 & \textbf{89.8} & 1.79\\ \hline
\textsc{[Norm]}  & 39.1 & 78.2 & 75.8 & 93.9 & 49.5 & 85.3 & 1.96\\ \hline
\textsc{[Bert]} & \textbf{53.5} & 85.4 & 78.4 & 93.9 & \textbf{62.4} & 89.5 & 2.12\\ \hline
\textsc{[Distil-Bert]}  & 49.8 & 85.8 & 76.6 & 93.7 & 59.3 & 89.6  & 2.17\\ \hline
\textsc{[RoBerta]} & 48.6 & 84.5 & 41.3 & 85.2 & 43.0 & 84.8 & 2.2\\ \hline\hline
\textbf{Method-1} &  36.0 & \textbf{89.4} & 36.4 & 88.0 & 36.2 & 88.6 & 1.83\\ \hline
\textbf{Method-2} & 12.1 & 35.5 & 63.3 & 95.0 & 18.6 & 51.4 &2.24\\ \hline\hline
\end{tabular}
\label{tab:concept-recognition-results}
\end{table}

\begin{figure*}
\centering
\includegraphics[width=0.925\textwidth]{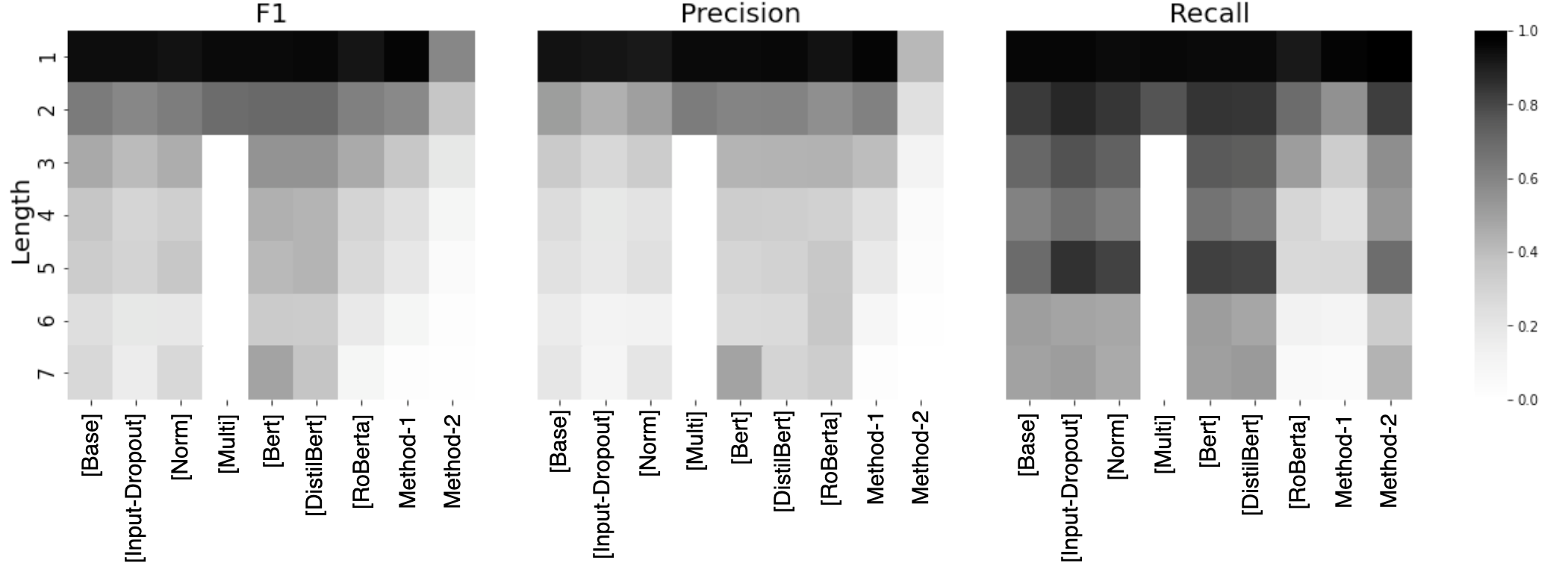}
\caption{Concept Recognition Performance by Word-Length}
\label{fig:concept-overview}
\end{figure*}

\subsubsection{Word-Length Performance}
\autoref{fig:concept-overview} shows the performance for every word-length, every model, and metric. 
This visualisation reveals additional insights and differences in the characteristic of the different models. 
Comparing the BERT-based models with the LSTM-based shows the reason for their higher macro scores.
They beat the LSTM-based on all levels for \textbf{F1}, and \textbf{P}.
An example is the performance of \textsc{[Bert]} on word-length seven, where it reaches an \textbf{F1} \& \textbf{P} of almost 50.0\%.
Further, all the proposed models, except of \textsc{[Multi]}, perform significant better than \textbf{Method-2} and \textbf{Method-1} for the length two to seven.
This gap explains the significant performance differences when considering the macro scores.
Besides, this figure clearly shows the reason for the low macro scores of \textsc{[Multi]} - it has a zero-performance for the lengths three to seven. 

\subsubsection{Nested-Level Performance}
The nested concepts' analysis reports the performance of the test set - grouped by the five nested-levels. 
\autoref{tab:concept-recognition-nested} shows the \textbf{F1} results for each level and each model. 
All the variations have a decreasing performance when the nested-level rise.
The \textsc{[Bert]} and \textsc{[DistilBert]} model can reduce this decreasement and have highest scores overall levels. 
\textsc{[Multi]} has a zero performance from level three since it does not predict any concept with three words or more.

\begin{table}
\centering
\caption{\textbf{F1} performance for all nested-levels for \textbf{CR}}
\begin{tabular}{@{\extracolsep{4pt}}>{\centering\arraybackslash} p{2cm} >{\centering\arraybackslash} p{0.5cm} >{\centering\arraybackslash}p{0.5cm} >{\centering\arraybackslash}p{0.5cm} >{\centering\arraybackslash}p{0.5cm} >{\centering\arraybackslash}p{0.5cm} }
\hline
&\multicolumn{5}{c}{\textbf{Nested Level}} \\ \cline{2-6} 

 & \textbf{1st}   & \textbf{2nd}   & \textbf{3rd} & \textbf{4th} & \textbf{5th}   \\ \hline
\hline
\textsc{[Base]} &  94.4 & 63.9 & 50.7 & 48.5 & 37.5 \\ \hline
\textsc{[Input-Drop]} & 93.6 & 59.0 & 44.0 & 41.8 & 22.7\\ \hline
\textsc{[Multi]} & \textbf{95.4} & 69.7 & 0.0 & 0.0 & 0.0\\ \hline
\textsc{[Norm]} & 92.9 & 63.2 & 50.0 & 47.7 & 25.6\\ \hline
\textsc{[Bert]} & 95.1 & \textbf{71.5} & 57.8 & \textbf{58.4} & \textbf{37.0}\\ \hline
\textsc{[DistilBert]} & 95.2 & 71.3 & \textbf{58.6} & 55.9 & 31.2\\ \hline
\textsc{[RoBerta]} & 91.1 & 62.0 & 45.5 & 24.7 & 22.2\\ \hline
\hline
\end{tabular}
\label{tab:concept-recognition-nested}
\end{table}

\subsection{Named Entity Recognition}
We report in the following section, the results of the different analyses for the NER experiment.
This comparison considers the overall, word-length, and nested-level performance by comparing the different variations and recent approaches (\textbf{BiFlaG} \cite{luo2020bipartite}, \textbf{BERT-MRC} \cite{li2019unified}, \textbf{SecondBest} \cite{shibuya2019nested}, \textbf{Layered-CRF} \cite{ju2018neural}).

\subsubsection{Overall Performance}

\autoref{tab:ner-overall} shows the overall performance of the different network variations.
It reveals that - in contrast to CR - LSTM-based models have higher performance (5.2\%-24.9\%) when it comes to \textbf{F1} and \textbf{P}, and at the same time, the BERT-based ones reach a higher \textbf{R} score (1.7\%-36\%).
For  \textbf{F1}, the \textsc{[Base]}, \textsc{[Norm-Flair]}, \textsc{[Norm]}, and \textsc{[Input-Drop]} have the highest performance, while \textsc{[Multi]}, a \textsc{[RoBerta]} have lowest one.

\begin{table}
\centering
\caption{Overall \textbf{NER} performance}
\begin{tabular}{>{\centering\arraybackslash}p{5cm}>{\centering\arraybackslash}p{0.6cm} >{\centering\arraybackslash}p{0.6cm} >{\centering\arraybackslash}p{0.6cm}}
\hline
\textbf{Model}  & \textbf{P} & \textbf{R} &\textbf{F1} \\ \hline \hline
\textsc{[Base]} &   58.2 & 79.2 & 67.1 \\ \hline
\textsc{[Input-Drop]} & 54.0 & 83.2 & 65.5\\ \hline
\textsc{[Multi]}  & 59.9 & 48.9 & 53.9\\ \hline
\textsc{[Norm]}  & 59.2 & 82.1 & 68.8\\ \hline
\textsc{[Norm-Flair]}  &\textbf{73.7} & 77.9 & \textbf{75.7}\\ \hline
\textsc{[Bert]} & 45.9 & 83.8 & 59.3\\ \hline
\textsc{[DistilBert]}  & 48.8 & \textbf{84.9} & 61.9\\ \hline
\textsc{[RoBerta]} & 34.1 & 81.0 & 48.0\\ \hline\hline
\textbf{BiFlaG}  \cite{luo2020bipartite}& 77.4 & 74.6 & 76.0\\ \hline
\textbf{BERT-MRC} \cite{li2019unified} &\textbf{85.2}& \textbf{81.1} & \textbf{83.8}\\ \hline
\textbf{SecondBest} \cite{shibuya2019nested}& 77.8 & 76.9 & 77.4\\ \hline
\textbf{Layered-CRF} \cite{ju2018neural}& 78.5 & 71.3 & 74.7\\ \hline\hline
\end{tabular}
\label{tab:ner-overall}
\end{table}

\begin{figure}
\centering
\includegraphics[width=0.5\textwidth]{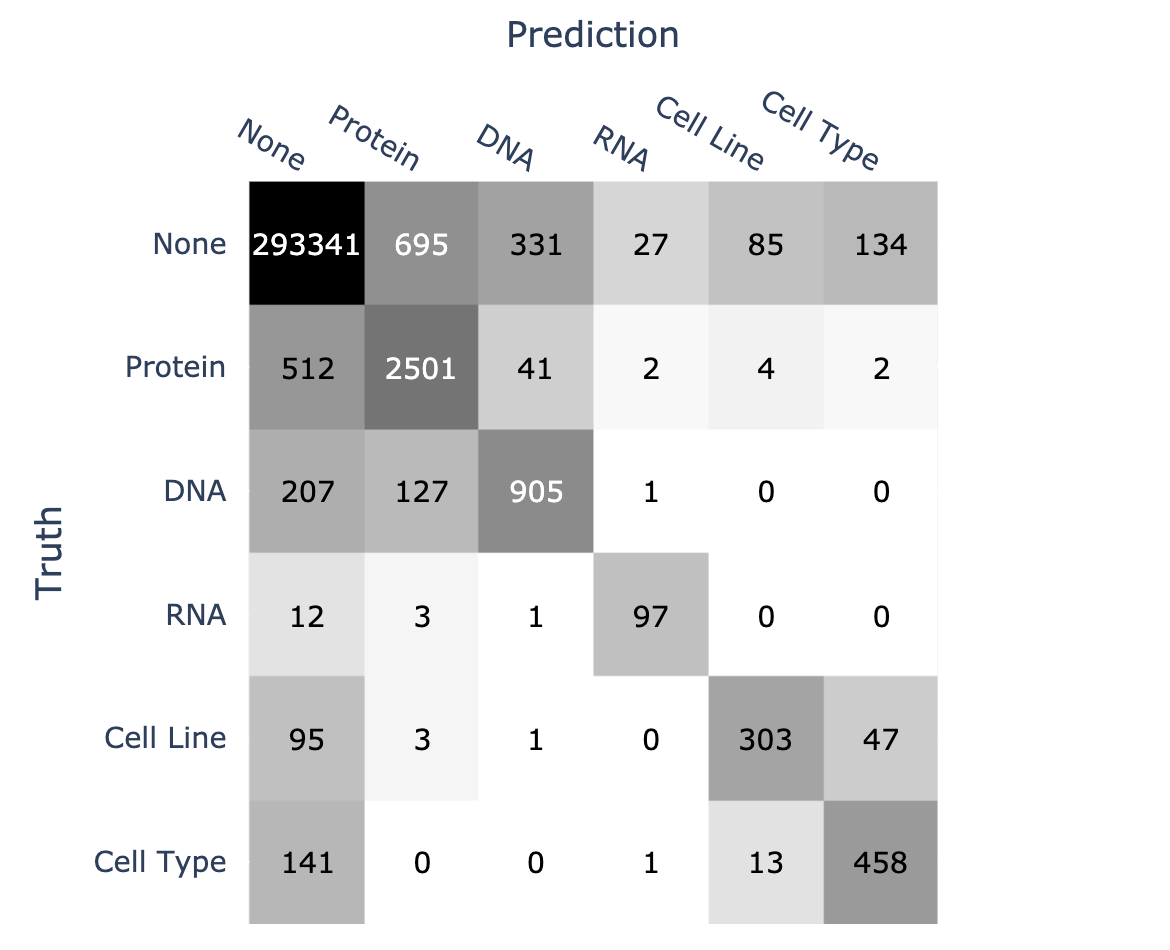}
\caption{Confusion matrix of the predictions of the \textsc{[Norm-Flair]} model.}
\label{table:ner-confusion-matrix}
\end{figure}

\autoref{table:ner-confusion-matrix} gives additional insights of the best-performing model \textsc{[Norm-Flair]} by showing the confusion matrix.
It shows the main reason for errors made lies in assigning a label to candidates that should not get any or in assign no label when required.
These two type cover 51\% respectively 39\% of all errors.
This behaviour is not pronounced for a specific label since this errors follows approximately the true distribution of the labels (as shown in \autoref{tab:portion-ner}).

The leftover 10\% of all errors are due to predicting the wrong label.
The most significant errors of this kind are predicting a \textbf{Protein} instead of \textbf{DNA} (10.5\% of the \textbf{DNA} cases) and predicting a \textbf{Cell Type} instead of \textbf{Cell Line} - 10.4\% of the \textbf{Cell Line} cases.

Compared to the four previous approaches, our models have higher \textbf{R} scores and lower \textbf{P}.
With regards to \textbf{F1}, the best performing model \textsc{[Norm-Flair]} outperforms \textbf{Layered-CRF} while underpeforming \textbf{BiFlaG}, \textbf{SecondBest}, and \textbf{BERT-MRC}.

The detailed results (as in \autoref{table:ner-per-tag}) compares our best variation \textsc{[Flair-Norm]} with two previous approaches \textbf{BiFlaG} \cite{luo2020bipartite} and \textbf{Layered-CRF} \cite{ju2018neural}.
Our model has a higher \textbf{R} (1.6-6.6\%), lower \textbf{P} (3.7-4.8\%), and 1\% higher \textbf{F1} than \textbf{Layered-CRF} and 0.3\% lower than \textbf{BiFlaG}.
The per-tag results show the tendency of a higher \textbf{R}, a lower \textbf{P}, and a comparable \textbf{F1}.

\begin{table*}
\centering
\begin{minipage}{.45\linewidth}
\caption{NER performance of \textbf{\textsc{[Norm-Flair]}} compared with \textbf{BiFlaG} \& \textbf{Layered-CRF}}
\begin{tabular}[t]{
>{\centering\arraybackslash}p{2.3cm}
>{\centering\arraybackslash}p{0.2cm}
>{\centering\arraybackslash}p{0.4cm}
>{\centering\arraybackslash}p{0.4cm}
>{\centering\arraybackslash}p{0.4cm}
>{\centering\arraybackslash}p{0.4cm}
>{\centering\arraybackslash}p{0.4cm}
>{\centering\arraybackslash}p{0.6cm}
}
\hline
    \\[16px]
     &  & \rotatebox{90}{\textbf{Protein}} & \rotatebox{90}{\textbf{DNA}}  & \rotatebox{90}{\textbf{RNA}}  & \rotatebox{90}{\textbf{Cell Line}} & \rotatebox{90}{\textbf{Cell Type}} & \rotatebox{90}{\textbf{Overall}\text{ }}  \\
\hline
\hline
\multirow{3}{*}{\textbf{\textsc{[Norm-Flair]}}}  & \textbf{P}      & 75.1    & 70.8 & 75.8 & 74.8      & 71.5      & 73.7     \\\cline{2-8} 
                             & \textbf{R}      & \textbf{81.7}    & \textbf{73.0}   & \textbf{85.8} & 67.5      & \textbf{74.7}      & \textbf{77.9}     \\ \cline{2-8} 
                             & \textbf{F1}     & 78.3    & 71.9 & 80.5 & 71.0       & 73.1      & 75.7     \\ \hline
\multirow{3}{*}{\textbf{BiFlaG}\text{ }\cite{luo2020bipartite}}      & \textbf{P}      & 79.5    & 72.7 & 84.4 & 75.9      & \textbf{76.7}      & 77.4     \\ \cline{2-8} 
                             & \textbf{R}      & 76.5    & 72.7 & 84.4 & \textbf{67.6}      & 72.4      & 74.6     \\ \cline{2-8} 
                             & \textbf{F1}     & \textbf{78.0}     & \textbf{72.7} & 84.4 & \textbf{71.5}      & \textbf{74.4}      & \textbf{76.0}       \\\hline
\multirow{3}{*}{\textbf{Layered-CRF}\text{ }\cite{ju2018neural}} & \textbf{P}      & \textbf{80.5}    & \textbf{74.4} & \textbf{90.3} & \textbf{77.8}      & 76.4      & \textbf{78.5}     \\ \cline{2-8} 
                             & \textbf{R}      & 73.2    & 69.7 & 79.5 & 65.7      & 68.1      & 71.3     \\ \cline{2-8} 
                             & \textbf{F1}     & 76.7    & 72.0   & \textbf{84.5} & 71.2      & 72.0        & 74.7  \\   \hline
                             \hline
\end{tabular}
\label{table:ner-per-tag}
\end{minipage}
\begin{minipage}{.48\linewidth}
\caption{Comparison of the nested \textbf{NER} performance on the GENIA dataset}
\begin{tabular}[t]{>{\centering\arraybackslash}p{0.15cm} >{\centering\arraybackslash}p{0.15cm} >{\centering\arraybackslash}p{0.2cm} >{\centering\arraybackslash}p{0.4cm} >{\centering\arraybackslash}p{0.4cm} >{\centering\arraybackslash}p{0.4cm} >{\centering\arraybackslash}p{0.4cm} >{\centering\arraybackslash}p{0.4cm} >{\centering\arraybackslash}p{0.4cm} >{\centering\arraybackslash}p{0.4cm} >{\centering\arraybackslash}p{0.4cm} >{\centering\arraybackslash}p{0.5cm}}
\hline
\multicolumn{1}{l}{}          &            &            & \rotatebox{90}{\textsc{[Base]}} & \rotatebox{90}{\textsc{[Input-Drop]}} & \rotatebox{90}{\textsc{[Multi]}} & \rotatebox{90}{\textsc{[Norm]}} & \rotatebox{90}{\textsc{[Norm-Flair]}} & \rotatebox{90}{\textsc{[Bert]}} & \rotatebox{90}{\textsc{[DistilBert]}} & \rotatebox{90}{\textsc{[RoBerta]}} & \rotatebox{90}{\textbf{SecondBest \text{ }\cite{shibuya2019nested}}} \\ \hline\hline
\multirow{9}{*}{\rotatebox{90}{\textbf{Nested Level}}} & \multirow{3}{*}{\rotatebox{90}{\textbf{1st}}} & \textbf{P}  & 60.0   & 55.3   & 60.8    & 61.5   & \textbf{77.1}             & 51.7   & 54.4   & 38.1 & -     \\ \cline{3-12} 
&                      & \textbf{R}  & 78.8   & 82.8   & 51.3    & 81.7   & 77.1   & 83.9   & \textbf{85.0}             & 82.1& 77.9   \\ \cline{3-12} 
&                      & \textbf{F1} & 68.1   & 66.3   & 55.7    & 70.2   & \textbf{77.1}             & 63.9   & 66.3   & 52.0& -      \\ \cline{2-12} 
& \multirow{3}{*}{\rotatebox{90}{\textbf{2nd}}} & \textbf{P}  & 49.0   & 47.2   & 45.8    & 47.3   & \textbf{58.3}             & 23.8   & 26.6   & 17.0& -      \\ \cline{3-12} 
&                      & \textbf{R}  & 83.9   & \textbf{86.9}             & 25.7    & \textbf{86.9}       & 85.5   & 82.7   & 83.8   & 69.5& 40.6   \\ \cline{3-12} 
&                      & \textbf{F1} & 61.9   & 61.2   & 32.9    & 61.3   & \textbf{69.3}             & 36.9   & 40.4   & 27.3& -      \\ \cline{2-12} 
& \multirow{3}{*}{\rotatebox{90}{\textbf{3rd}}} & \textbf{P}  & 12.1   & \textbf{15.0}             & 0.0     & 12.9   & 14.0   & 6.5    & 6.4    & 4.7 & -      \\ \cline{3-12} 
&                      & \textbf{R}  & 68.8   & \textbf{93.8}             & 0.0     & 81.2   & 81.2   & 93.3   & 86.7   & 80.0& -      \\ \cline{3-12} 
&                      & \textbf{F1} & 20.6   & \textbf{25.9}             & 0.0     & 22.2   & 23.9   & 12.2   & 11.9   & 8.9 & -      \\ \hline\hline
\end{tabular}
\label{table:ner-nested}
\end{minipage}
\end{table*}

\subsubsection{Word-Length Performance}
\begin{figure*}
\centering
\includegraphics[width=1\textwidth]{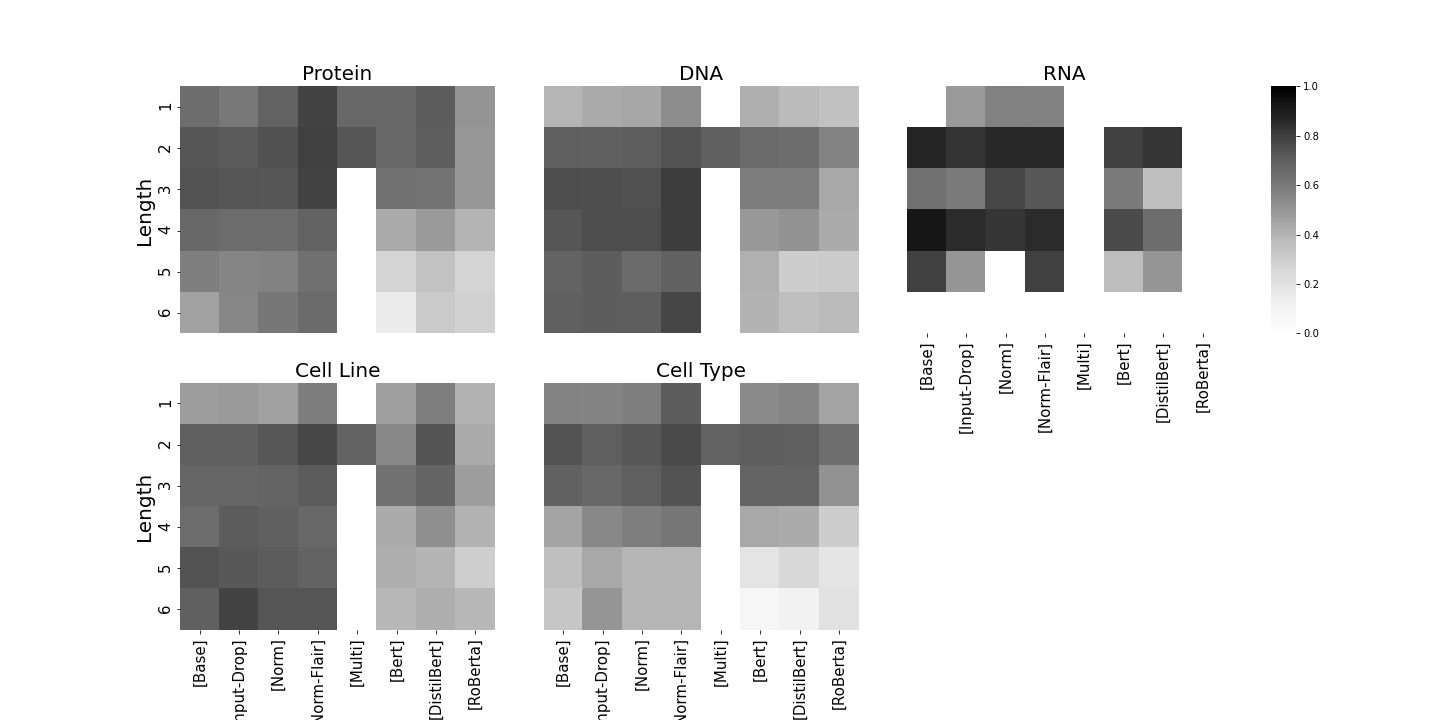}
\caption{Overview of the \textbf{F1} for the different models, levels, and labels}
\label{fig:ner-overview}
\end{figure*}

\autoref{fig:ner-overview} gives an overview of the \textbf{F1} performance with regards to every label, level, and model.
It shows that the LSTM-based models have a more balanced performance than the BERT-based ones.
One can also see that the models have their highest performance on the second level.
The \textsc{[Multi]} model has - similar to the previous experiment - a zero performance for levels three to six, and the level one except for the \textbf{Protein} label.
Further, we see a sparse performance for the \textbf{RNA} label.
While some models reached a high score on level two, four, and five, they all scored below average for level one.
\textsc{[RoBerta]} and \textsc{[Multi]} missed any sample of this label and have a zero performance for all levels.

\subsubsection{Nested-Level Performance} 
\autoref{table:ner-nested} shows the detailed performance of the different model variations per nested-level.
Overall all of them have the highest scores for the 1st level, slightly lower one on the second level, and an apparent decrease for 3rd level.
As in the previous results, the variation \textsc{[Norm-Flair]} has the best performance.
It has the highest \textbf{P} and \textbf{F1} for the 1st level (77.1\%, 77.1\%), and for the 2nd one (58.3\%, 69.3\%).
\textsc{[Input-Drop]} has the highest \textbf{R} score for level two (86.9\%) and the best \textbf{P}, \textbf{R}, and \textbf{F1} on the 3rd level (15.0\%, 93.8\%, 25.9\%)
The LSTM-based models achieved better results than the BERT-based ones, except \textsc{[DistilBert]} that reached the highest \textbf{R} one the 1st level. 
Further, our approach outperforms \textbf{SecondBest}, the only previous work that reports \textbf{R} for all nested-levels, for level one (0.8\%-7\%) and two (29.5\%-46.9\%).

\subsection{Case Study}
This case study focus on both tasks - CR and NER - separately. 
It analyses the prediction of the various models based on a given sentence. 
This qualitative analysis reveals the differences between those models, where they are successful and where they fail.

\subsubsection{Concept Recognition}

\autoref{table:concept-samples} shows the recognised concepts, within the example sentence, for all the considered models.
This qualitative evaluation reflects the quantitative results shown in \autoref{tab:concept-recognition-results}:
\begin{itemize}
    \item Models with a high precision tends to recognise fewer false-positive concepts. However, at the same time, it misses some of the concepts (false negatives). 
    As an example, \textsc{[Bert]} just recognizes one concept wrongly - \textit{Naay}. 
    Thereby, it reaches a precision of 83.3\% and simultaneously a recall of 83.3\% by missing the concept \textit{Council of the Federation}.
    \item In contrast, models that gain an overall higher recall catch more true concepts.
    However, they mistakenly recognise concepts more often.
    Exemplary, the models \textsc{[Base]} and \textsc{[Input-Drop]} reach a recall of 100\% by including the concept \textit{Council of the Federation}, but likewise a precision of 66.7\% resp. 46.2\% due to the wrongly recognition of non-concepts like or \textit{XaaydagGa} or \textit{Council of the Federation of Haida}.
\end{itemize}

Taking a closer look at the wrongly recognised reveals, one can recognise them as valid concepts for formal and semantic reasons. 
First of all, they fulfil the formal requirements of a concept following the formal requirements (defined in \cite{parameswaran2010towards}).
They all contain at least one nouns and do not start and end with a verb, conjunction, article or pronoun.
Further, from a semantic point of view, \textit{XaaydaGa Waadluxan Naay} is the native name of the government of \textit{Haida Nation}. By that, it is a crucial part of the sentence from an information perspective.
Thereby, one can see candidates like \textit{XaaydaGa Waadluxan Naay} or \textit{Council of the Federation of Haida} as valid concepts even.

\setlength\tabcolsep{1.1pt}
\begin{table}
\centering
\caption{Detailed \textbf{CR} predictions for the sentence \textit{XaaydaGa Waadluxan Naay is the elected government and the Council of the Federation of Haida}. True positives are marked with a uppercase \textbf{C} and false positives with a lowercase \textbf{c}.}
\begin{tabular}{@{\extracolsep{1pt}}>{\centering\arraybackslash} p{2.1cm} c c c c c c c c c c c c c c c c c }
\hline
\textbf{Model} & 
\rotatebox{90}{\textit{Council}}&
\rotatebox{90}{\textit{elected}}& 
\rotatebox{90}{\textit{Federation}}& 
\rotatebox{90}{\textit{government}}& 
\rotatebox{90}{\textit{Haida}}& 
\rotatebox{90}{\textit{Naay}}& 
\rotatebox{90}{\textit{Waadluxan}}&
\rotatebox{90}{\textit{XaaydagGa}}&
\rotatebox{90}{\textit{elected government}}& 
\rotatebox{90}{\textit{Waadluxan Naay}}&
\rotatebox{90}{\textit{XaaydagGa Waadluxa}}& \rotatebox{90}{\textit{Federation of Haida}}&
\rotatebox{90}{\textit{XaaydagGa Waadluxan Naay}}& \rotatebox{90}{\textit{Council of the Federation}}&  \rotatebox{90}{\textit{Council of the Federation of Haida}}
\\ \hline\hline
\textsc{[Base]} &
$ C$ &   & $ C$ & $ C$ & $ C$ & $ c$ & $ c$ & $ c$ & $ C$ &   &   &   &   & $ C$ & \\ \hline
\textsc{[Input-Drop]} &
$ C$ &   & $ C$ & $ C$ & $ C$ & $ c$ & $ c$ & $ c$ & $ C$ & $ c$ & $ c$ &   & $ c$ & $ C$ &  $ c$\\ \hline
\textsc{[Multi]}&
$ C$ &   & $ C$ & $ C$ & $ C$ & $ c$ & $ c$ &   & $ C$ & $ c$ &   &   &   &   &    \\ \hline
\textsc{[Norm]}&
$ C$ &   & $ C$ & $ C$ & $ C$ &   &   &   & $ C$ &   &   & $ c$ &   &    & $ c$\\ \hline
\textsc{[Bert]}&
$ C$ &   & $ C$ & $ C$ & $ C$ & $ c$ &   &   & $ C$ &   &   &   &   &   &    \\ \hline
\textsc{[DistilBert]}&
$ C$ &   & $ C$ & $ C$ & $ C$ & $ c$ & $ c$ &   & $ C$ &   &   & $ c$ &   &   &   \\ \hline
\textsc{[RoBerta]}&
$ C$ &   & $ C$ & $ C$ & $ C$ & $ c$ & $ c$ & $ c$ & $ C$ &   &   &   &   &   &   \\ \hline\hline
\end{tabular}
\label{table:concept-samples}
\end{table}

\subsubsection{Named Entity Recognition}

The \autoref{table:ner-samples} shows the detailed prediction-analysis of the different models. 
These results confirm the overall results (show in \autoref{tab:ner-overall}):
\begin{itemize}
    \item The \textsc{[Norm-Flair]} model, by far the one with the highest overall precision, is the only one with a precision of 100\%. 
    But on the same time it misses three samples (\textit{TCR}, \textit{TCR beta}, and \textit{enhancers}) and gains a recall of 57\%. 
    \item On the contrary, \textsc{[Input-Drop]} achieves a recall of 86\% by just missing the true sample \textit{TCR}.
    However, it has a recall 86\% due to the assignment of the \textit{DNA} label to the sample \textit{5'-GGCACCCTTTGA-3'}.
    \item The bert-based models (\textsc{[Bert]}, \textsc{[DistilBert]}, and \textsc{[RoBe\-rta]}) have a low precision of 25\% to 50\% - as in the overall evaluation. 
    In contrast to the overall results, they have a low recall of 14\% - 28\% for this sample sentence.
\end{itemize}

\begin{table}
\centering
\caption{Detailed NER predictions for the sentence \textit{Sequences related to the TCF-1 alpha binding motif (5'-GGCACCCTTTGA-3') are also found in the human TCR delta (and possibly TCR beta) enhancers}. True positives are marked with a uppercase letter and false positives with a lowercase one - P for Protein and D for DNA.}
\begin{tabular}{@{\extracolsep{1pt}}>{\centering\arraybackslash} p{2.2cm} c c c c c c c c c c c c }
\hline
\textbf{Models}& \rotatebox{90}{\textit{TCF-1 alpha}} & \rotatebox{90}{\textit{TCF-1 alpha binding motif}} & \rotatebox{90}{\textit{human TCR delta}} & \rotatebox{90}{\textit{TCR}} & \rotatebox{90}{\textit{TCR delta}} & \rotatebox{90}{\textit{TCR beta}} & \rotatebox{90}{\textit{enhancers}} & \rotatebox{90}{\textit{TCF-1}} & \rotatebox{90}{\textit{TCF-1 alpha binding}} & \rotatebox{90}{\textit{5'-GGCACCCTTTGA-3'}} & \rotatebox{90}{\textit{delta}} & \rotatebox{90}{\textit{beta}}\\ \hline \hline
\textsc{[Base]} & P & D & p & - & - & D & - & - & - & d  & - & - \\ \hline
\textsc{[Input-Drop]} & P  & D  & D   & - & D & D & D & - & - & d & - & -\\ \hline
\textsc{[Multi]}& P  & -  & -   & - & -& p  & -& - & - & - & - & -\\ \hline
\textsc{[Norm]} & P  & D  & p   & - & D & D & D & - & - & d & - & -\\ \hline
\textsc{[Norm-Flair]}  & P  & D  & D   & - & D & -   & -& - & - & - & - & -\\ \hline
\textsc{[Bert]} & P  & D  & -   & - & -& -   & -& p & - & d & - & -\\ \hline
\textsc{[DistilBert]}  & P  & -  & -   & - & -& - & -& p & - & d & - & -\\  \hline
\textsc{[RoBerta]}  & P  & D  & - & - & - & p & -& p & d & d & p & p\\ \hline\hline
\end{tabular}
\label{table:ner-samples}
\end{table}

\section{Discussion}
With this paper, we present an approach for nested entities recognition.
It uses a partly-layered network architecture that handles nested and overlapping entities of varying lengths.
We evaluate eight variations of this architecture on CR and NER tasks.
Every instance consists of either an LSTM- or a BERT-based ground layer and a selection of dense-, dropout-, and normalisation layers. 
The evaluation results show that we outperform previous CR approaches for word-length, nested-level, and overall performances.
Further, the concluding case study gives a feeling of the ability to generalise and handle unknown words. 
Regarding NER tasks, our approach achieves state-of-the-art results for precision, recall, F1 score.
Besides, the results demonstrate an adequate capacity of this architecture to generalise for all word-lengths.
Our approach also presents better recall when looking to the nested-level aspect than recent comparable work.

As one can see in both evaluated tasks, our proposed network architecture favours the recall measure. 
Additionally, this partly-layered network structure has the advantage of balancing the performances for all word-lengths considered.
Despite preventing the network from learning sharp decision boundaries, this structure helps regulate the model and handle the imbalance between different word-lengths. 
This behaviour is particularly evident by comparing a fully-layered variation such as \textsc{[Multi]} with the others, that are all partly-layered.
While reaching high-performance scores for word-length one and two, \textsc{[Multi]} does not generalise for higher lengths, due to the missing common structure. 
This fact implies a less effective learning-process for a word-length of three or higher.
To conclude, both experiments show our approach's effectiveness in terms of performances and the ability to fight the intrinsic imbalance within the different word-lengths.

The following aspects are on our agenda.
Comparing the current architecture (a Tagging-Layer per word-length) with an adjusted one - using a Tagging Layer per nested-level.
This structural modification will require on top the addition of a Conditional Random Field (CRF) layer and the usage of another target encoding schema like BIO (begin-inner-outer).
Despite these modifications, it could improve the performance for higher nested levels and foster applicability for different use cases, such as extraction of events and argumentation components within the text.
Such applications will allow us to combine layers for different tasks towards a more robust network and more stable results.

\bibliographystyle{ACM-Reference-Format}
\bibliography{refs}

\appendix

\end{document}